\documentclass[lettersize,journal]{IEEEtran}

\usepackage{amsmath,cleveref}
\usepackage{multirow}
\usepackage{amsmath,amsfonts}
\usepackage{graphicx,caption}
\usepackage{subcaption}
\usepackage{algorithmic}
\usepackage{array}
\usepackage{bbding}
\usepackage[caption=false,font=normalsize,labelfont=sf,textfont=sf]{subfig}
\usepackage{textcomp}
\usepackage{stfloats}
\usepackage{url}
\usepackage{verbatim}
\usepackage{graphicx}
\def\BibTeX{{\rm B\kern-.05em{\sc i\kern-.025em b}\kern-.08em
    T\kern-.1667em\lower.7ex\hbox{E}\kern-.125emX}}
\usepackage{balance}
\begin{document}
\title{RecipeGen: A Benchmark for Real-World Recipe Image Generation}
\author{Ruoxuan Zhang, Hongxia Xie, Yi Yao, Jian-Yu Jiang-Lin, Bin Wen, Ling Lo, Hong-Han Shuai, Yung-Hui Li, Wen-Huang Cheng}

\markboth{Journal of \LaTeX\ Class Files,~Vol.~18, No.~9, September~2020}%
{How to Use the IEEEtran \LaTeX \ Templates}

\maketitle

\begin{abstract}
Recipe image generation is an important challenge in food computing, with applications from culinary education to interactive recipe platforms. However, there is currently no real-world dataset that comprehensively connects recipe goals, sequential steps, and corresponding images. To address this, we introduce RecipeGen, the first real-world goal-step-image benchmark for recipe generation, featuring diverse ingredients, varied recipe steps, multiple cooking styles, and a broad collection of food categories. Data is in \url{https://github.com/zhangdaxia22/RecipeGen}.

\end{abstract}


\section{Introduction}
\IEEEPARstart{G}{enerating} illustrated instructions has gained growing attention as a means to assist users in comprehending and executing complex, step-by-step tasks. Whether assembling furniture, repairing devices, or following cooking procedures, visual depictions of each step can significantly reduce confusion and errors. One of the compelling applications in this space is recipe generation, where step-by-step illustrations enhance users’ understanding of cooking processes. Existing studies in recipe image generation can be categorized into creating step images from the recipe title and ingredients~\cite{h2020recipegpt}, from recipe steps~\cite{pan2020chefgan,liu2023ml,han2020cookgan} and ingredients, or from a reference dish image~\cite{chhikara2024fire,dosovitskiy2020image,wang2020structure,wang2022learning}. While these approaches have shown promise, they often struggle to generate intermediate steps with clarity and consistency, which leads to confusion and misinterpretation. A primary reason for this limitation lies in existing datasets~\cite{marin2021recipe1m+, bien-etal-2020-recipenlg,batra2020recipedb}, which typically only include final images of the dish, lacking detailed snapshots that capture essential transitions between steps.

To address these shortcomings, we introduce \textbf{RecipeGen}, a real-world goal-step-image benchmark tailored for generating illustrative step-by-step instructions in the cooking domain. We collect recipes spanning diverse cuisines, cooking methods, and food types, guided by 158 targeted keywords to ensure broad coverage. Each sample includes a dish name, ingredient list, detailed cooking steps, and corresponding images for every step. Altogether, our dataset comprises 21,944 recipes, totaling 139,872 images paired with textual descriptions. By retaining comprehensive visual details at each stage of food preparation, RecipeGen lays the groundwork for more accurate and instructive recipe-generation models, serving as a robust resource for the larger domain of generating illustrated instructions.

\section{Related Work}

Food is closely related to people's lives, and with the development of human society, diets have diversified. To facilitate the management of human life and health, the field of food computing  \cite{min2019survey} has emerged. Academically, topics such as  food segmentation \cite{lan2023foodsam,yin2023foodlmm}, food recognition \cite{min2023large,8779586,zhang2023deep}, food recommendation \cite{min2019food,wang2021market2dish}, food reasoning \cite{zhou2024foodsky} and recipe image generation have gained significant attention.

\textbf{Food Computing Dataset.}
There are several existing datasets for food computing tasks. For instance, Food-101 \cite{bossard2014food}, VireoFood-172 \cite{chen2016deep}, Recipes242k \cite{rokicki2018impact}. However, these datasets are primarily used for fine-grained tasks such as food or ingredients classification, nutritional analysis, calorie calculation, etc. In addition to these information, FoodEarth \cite{zhou2024foodsky} utilizes LLMs to develop a comprehensive dataset for food and nutrition analysis. Similarly, Recipe1M \cite{marin2021recipe1m+} provides recipe instructions linked to each final dish image. However, neither dataset includes images for each individual recipe step. While the recipe section of VGSI \cite{zhou2018towards} offers step images for each instruction, many of these images are presented in a comic style. To address this gap, we propose the RecipeGen Benchmark, which is the first dataset that provides a real-world step-image recipe dataset. 

\section{RecipeGen Benchmark}
\label{sec:benchmark}

\begin{figure*}[t]
    \centering
    
    \begin{subfigure}[b]{0.48\linewidth}
        \centering
        \includegraphics[width=\linewidth]{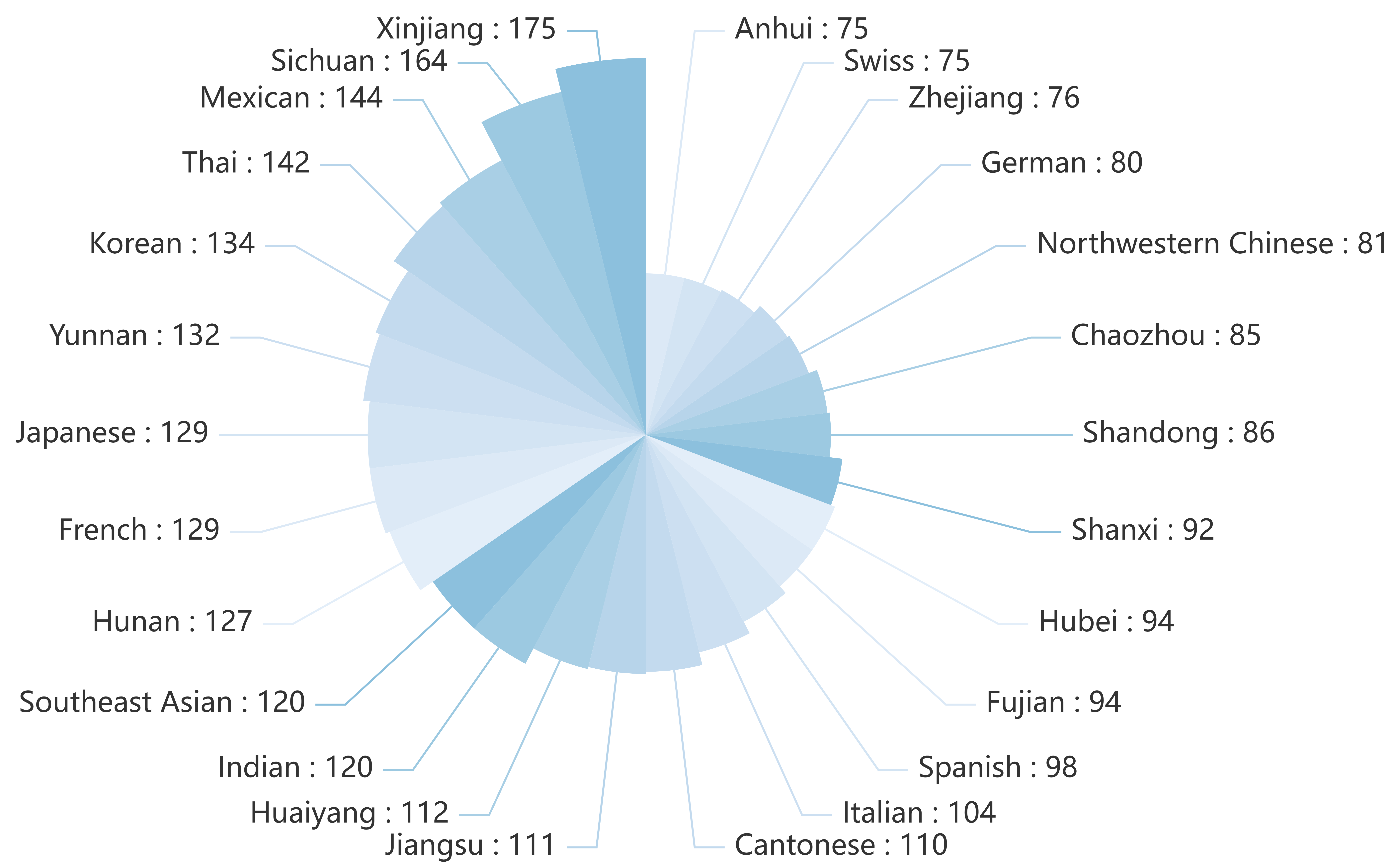}
        \caption{The distribution of region keywords.}
        \label{fig:image_a}
    \end{subfigure}
    \hfill
    \begin{subfigure}[b]{0.48\linewidth}
        \centering
        \includegraphics[width=\linewidth]{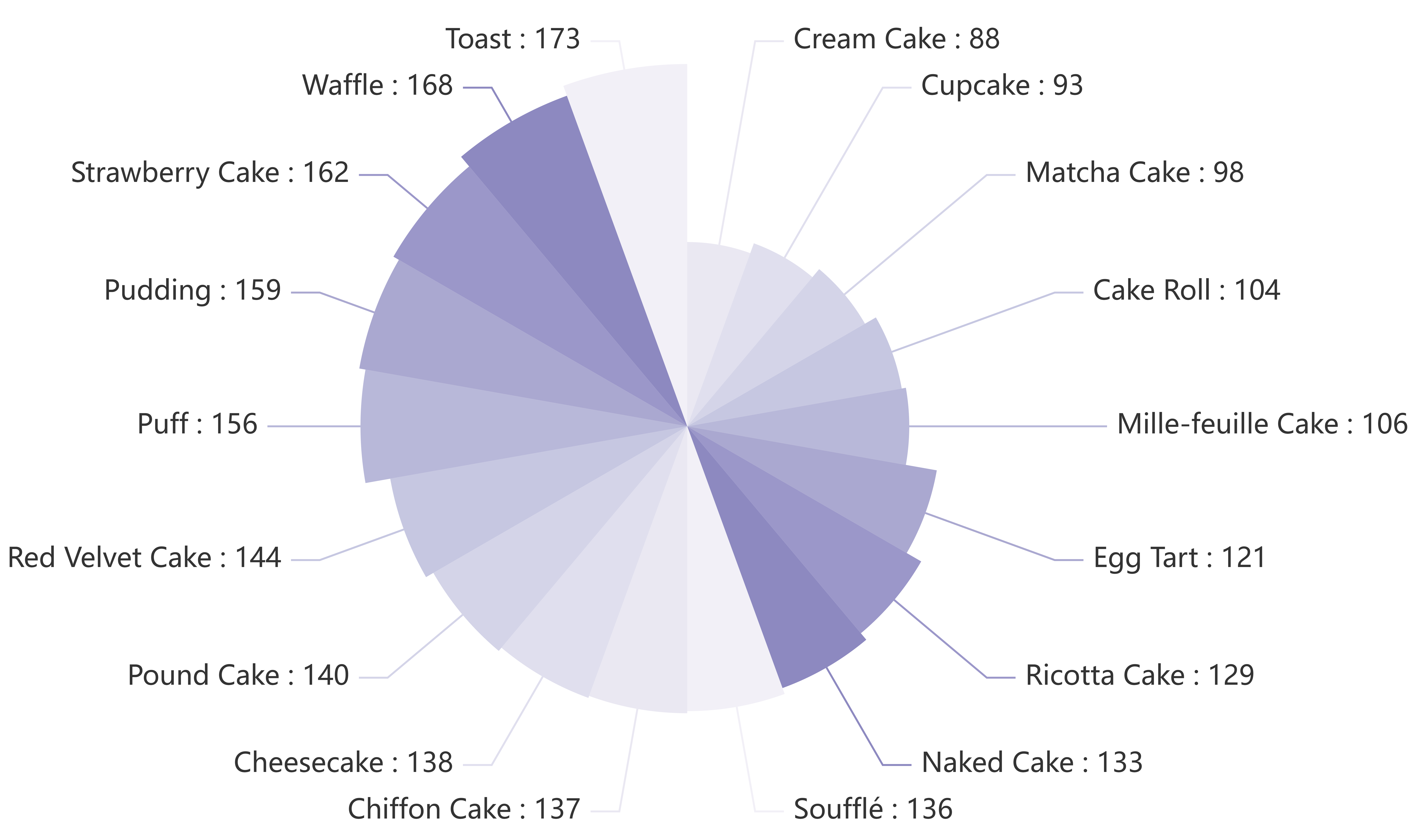}
        \caption{The distribution of dessert keywords.}
        \label{fig:image_b}
    \end{subfigure}
    \hfill
    \begin{subfigure}[b]{0.45\linewidth}
        \centering
        \includegraphics[width=\linewidth]{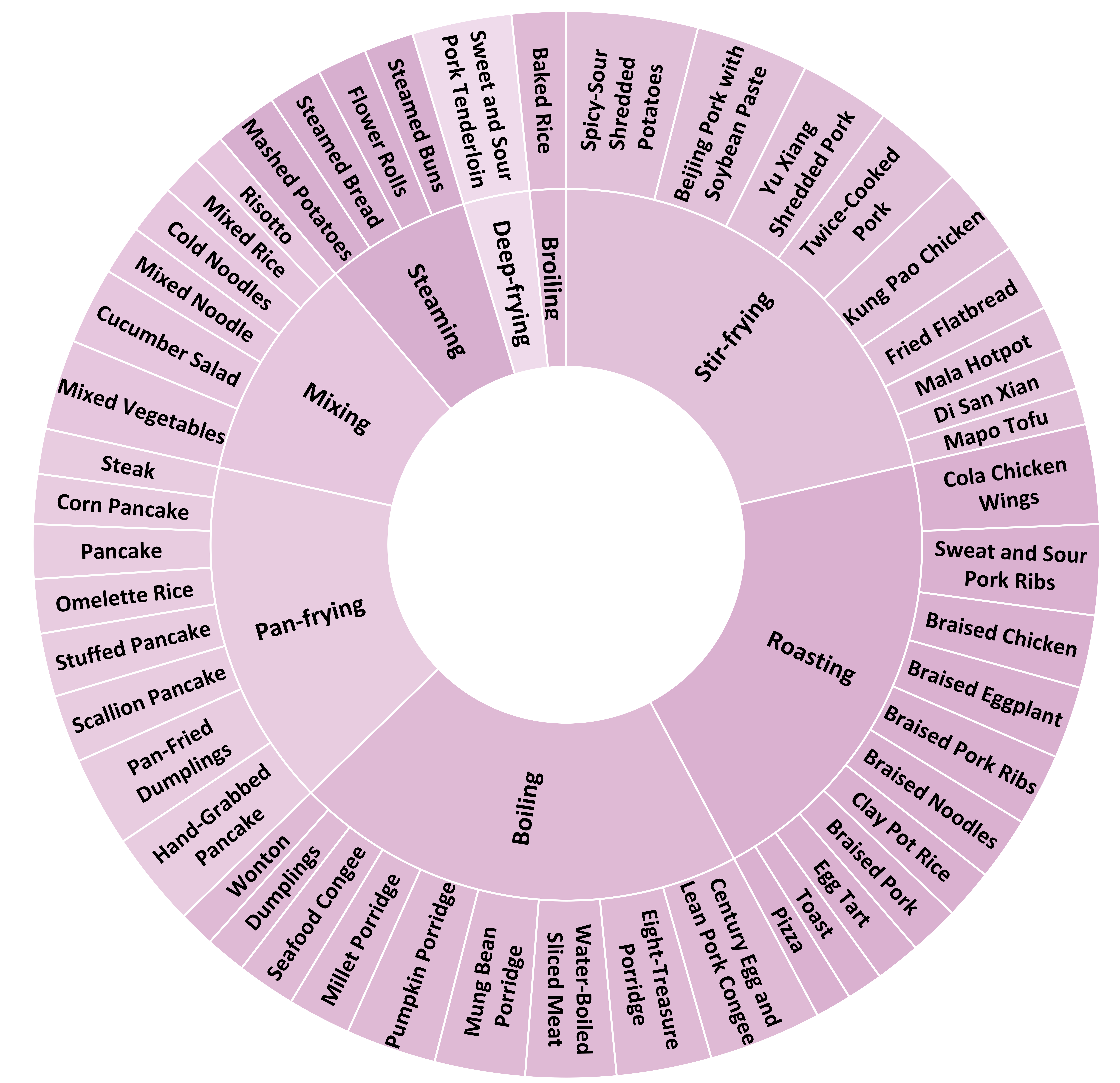}
        \caption{The distribution of cooking style keywords.}
        \label{fig:image_a}
    \end{subfigure}
    \hfill
    \begin{subfigure}[b]{0.45\linewidth}
        \centering
        \includegraphics[width=\linewidth]{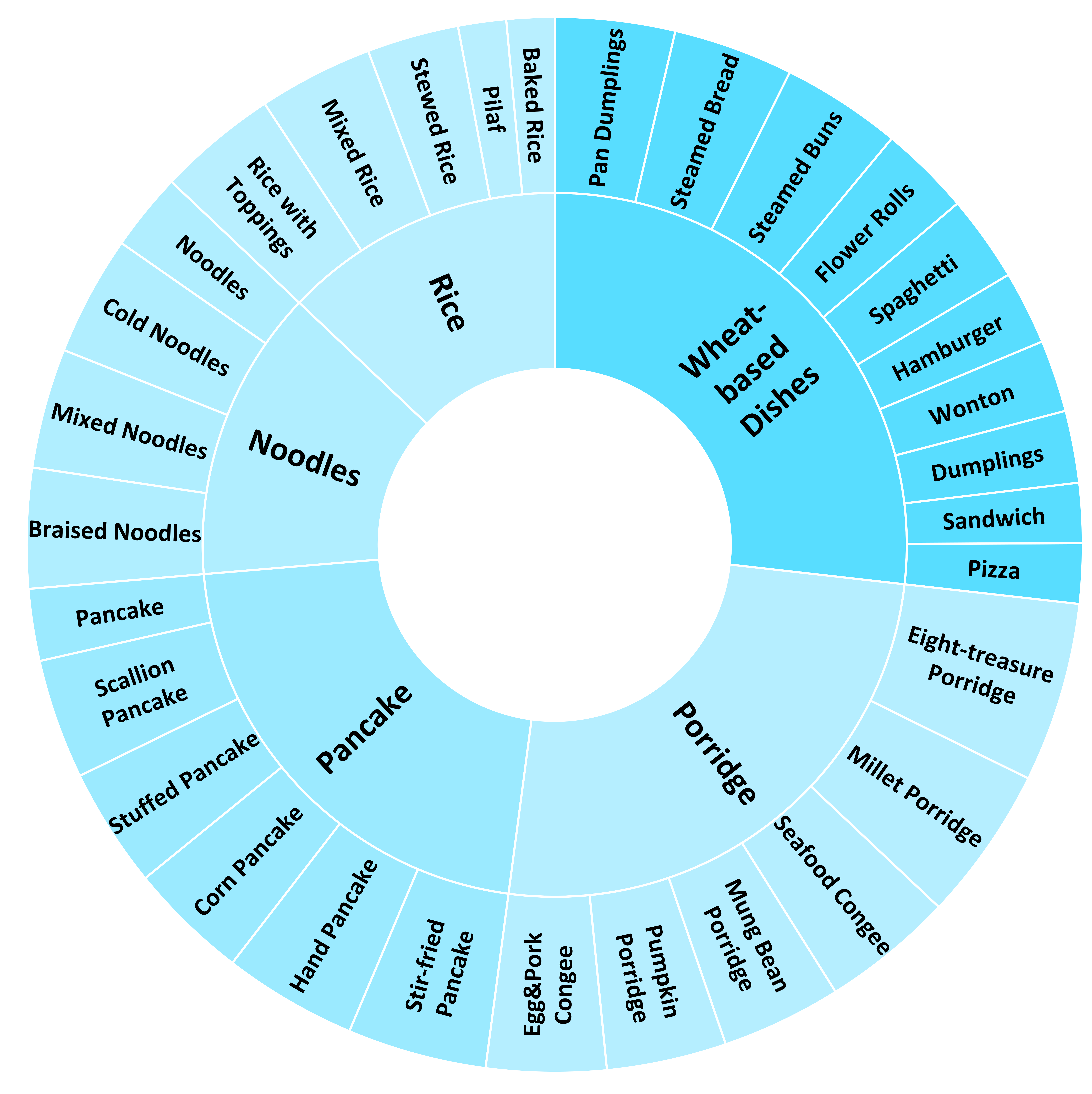}
        \caption{The distribution of staple keywords.}
        \label{fig:image_b}
    \end{subfigure}
    \hfill
    
    \caption{The distribution of region and dessert keywords in RecipeGen Benchmark.}
    \label{fig:data_region}
\end{figure*}

\begin{figure*}[t]
  \centering
 
  \includegraphics[width=1\linewidth]{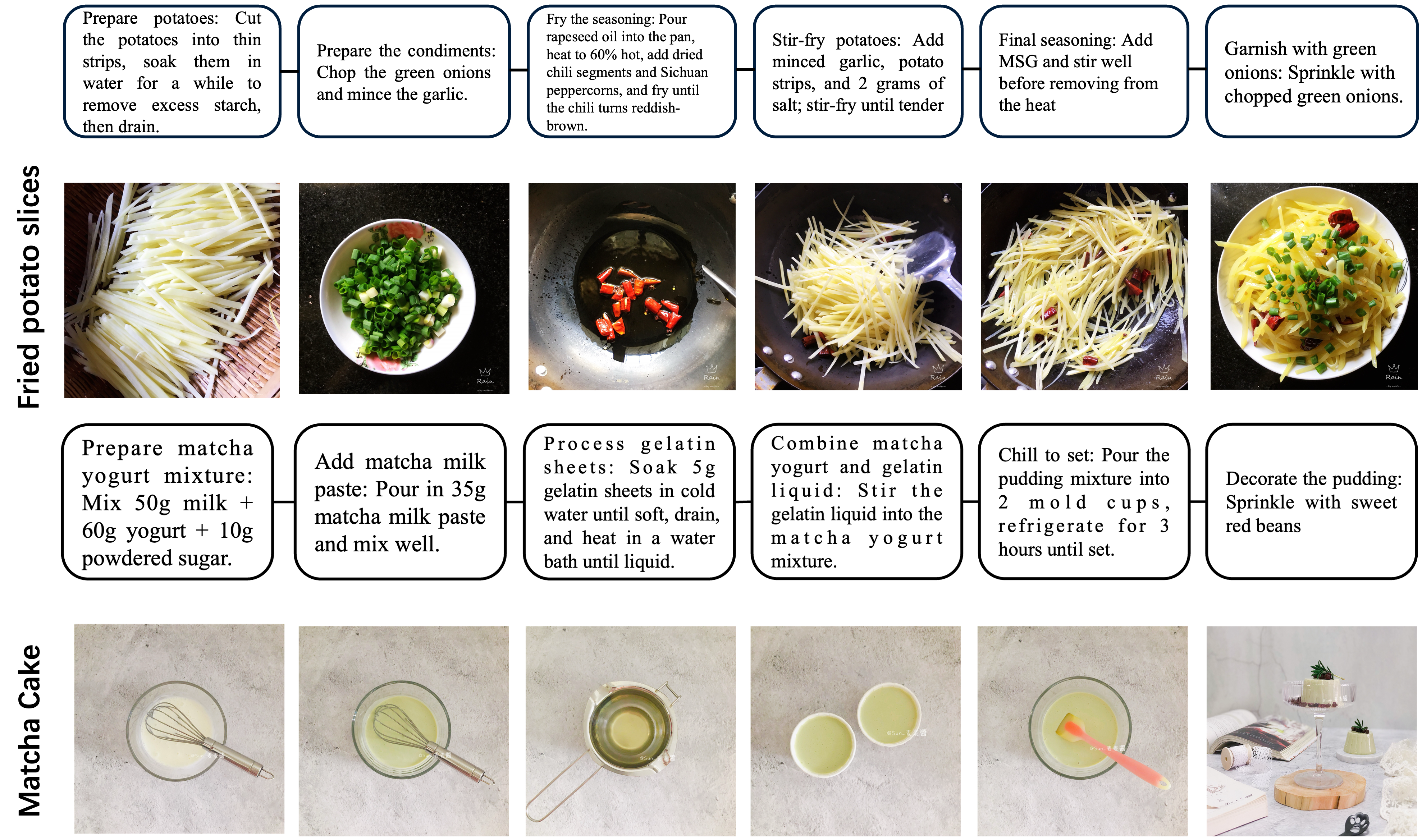} 

  \caption{Some examples in RecipeGen. Each sample contains its goal, steps, and the images corresponding to each step.
}
  \label{fig:example}
\end{figure*}

\begin{table*}[h!]
  \centering
  \tabcolsep=4.5pt
  \fontsize{7}{11}\selectfont
  \begin{tabular}{l r r r r r r r r r r r r} 
    \hline
    \multirow{2}{*}{DataSet} & \multirow{2}{*}{Categories} & \multirow{2}{*}{Recipes} & \multirow{2}{*}{Steps/Images} & \multirow{2}{*}{Modality} & \multirow{2}{*}{Source} & \multicolumn{2}{c}{Keywords}&\multicolumn{2}{c}{Data Faithfulness} & \multicolumn{3}{c}{Step Counts} \\ 
    \cline{7-8} \cline{9-10}\cline{11-13}
          &          &          &        &          &       &Chinese&Western& GF & SF & 0-6& 7-10&  \textgreater 10\\ 
    \hline
    VGSI-Recipe & - &1,157&6,417& Image & wikiHow &\XSolidBrush&\Checkmark& 81.40 & 73.70&72.95\%&24.37\%&2.68\%\\ 
    YouCook2 & 89 & 2,000 & 15,433 & Video & YouTube &\XSolidBrush&\Checkmark& - & - & 38.50\%&44.15\% &17.35\% \\ 
    \textbf{RGB (Ours)} & \textbf{158} & \textbf{21,944} & \textbf{139,872} & \textbf{Image} & \textbf{douguo} &\textbf{56+64}&\textbf{37+64}& \textbf{85.40} & \textbf{ 84.44} & \textbf{44.16\%}& \textbf{53.78\%}& \textbf{2.06\%} \\ 
    \hline
  \end{tabular}
  \caption{Comparison among Different Recipe Datasets. In the ``keywords" column of the table, it indicates whether the keyword category is Chinese cuisine or Western cuisine. The number before the plus sign represents the total images of keywords that exclusively belong to either Chinese or Western cuisine, while the number after the plus sign represents the count of keywords that are common to both cuisines.}
  \label{tab:RecipeDataset}
\end{table*}
The diversity of human culinary practices, developed over centuries, has led to a vast array of ingredient combinations and cooking methods, posing challenges for constructing comprehensive recipe step-image datasets. Existing datasets are limited, often constrained to narrow step counts and a restricted range of ingredients and interactions.

To address this gap, we propose \textbf{RecipeGen Benchmark (RGB)}, a goal-instruction image dataset designed to evaluate text-to-image models (T2Is) in understanding cooking instructions. RGB includes 21,944 recipes with 139,872 images and steps, encompassing diverse regional dishes and cooking styles. Collected from a large set of user-uploaded, real-world recipes\footnote{\url{www.douguo.com}}, RGB provides a solid foundation for the evaluation of the T2I model. This dataset captures the distinctive ways different culinary traditions prepare similar ingredients and combines unique cooking techniques. For instance, Western recipes may pan-fry beef as a steak, while East Asian cuisine might braise it with root vegetables. Unlike existing datasets focused primarily on Western cuisines, such as VGSI-Recipe \cite{yang2021visual}, RGB spans a wide range of regional styles, including East and Southeast Asian cuisines, enhancing model understanding of varied preparation techniques.

\subsection{Dataset Construction Procedure}

To ensure RGB’s diversity, we curated 158 keywords across different cuisines (e.g., Cantonese, Mexican), dish types (e.g., main courses, desserts), and cooking techniques (e.g., stir-frying, roasting), gathering 29,026 recipes.   

\begin{figure}[t]
  \centering
 
  \includegraphics[width=1\linewidth]{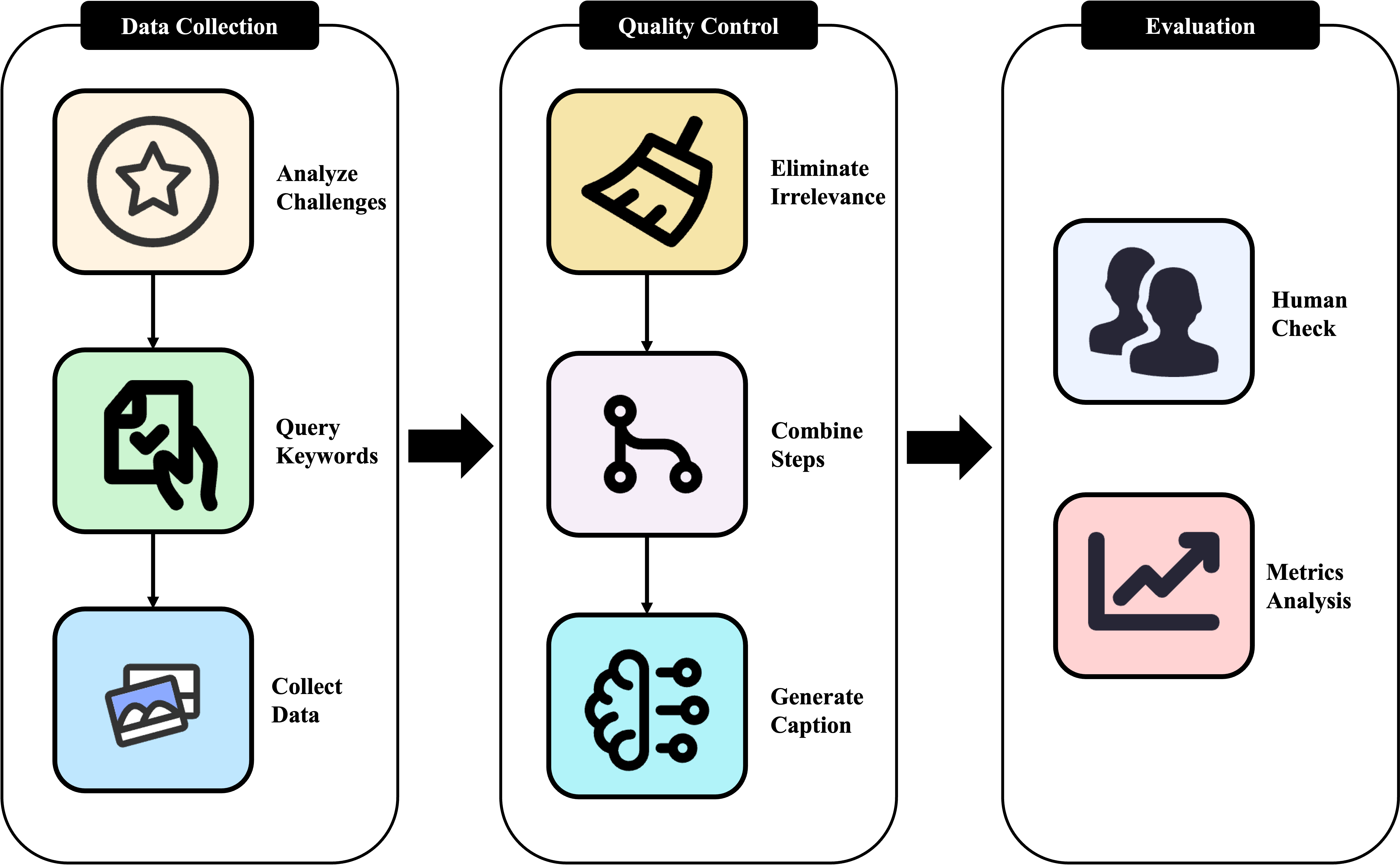} 

  \caption{Dataset Construction Procedure. We first analyze the characteristics of the dishes and select 158 keywords. Subsequently, we utilize GPT-4o to perform quality control by omitting irrelevant steps, merging adjacent simple actions, and generating captions. Finally, we calculate metrics and conduct human checks to ensure the usability of the dataset.
}
  \label{fig:gpt-benc}
\end{figure}

 \textbf{Quality Control.}
The process is shown in Fig. \ref{fig:gpt-benc}. To maintain consistency and clarity, we implement a quality control process to address common issues like incomplete instructions or overly detailed steps. This process involved: (1) Removing recipes with low quality (including lack images or steps, mismatched numbers of images and steps, vague directions \footnote{For example, some steps in the recipe only contain phrases like ``see as images" without providing clear or actionable instructions.}). In this step, we delete 4,978 recipes; (2) Using GPT-4o \cite{achiam2023gpt} to clean and refine recipes by eliminating irrelevant content \footnote{For instance, some users include exclamatory phrases such as ``This tastes amazing!" or suggestions like ``Using ingredient A instead of B works better."}, merging redundant steps, and summarizing for coherence. And then, we remove recipes where the LLM's hallucination issues caused a single step to be split into two, as well as recipes with incorrect output formats. A total of 2,104 recipes were deleted. After Quality Control, the number of recipe is 21,944 and the number of images is 139,872; (3) Computing data faithfulness and using human check to insure the quality. The result of metrics is in Tab. \ref{tab:RecipeDataset}. 

\begin{figure*}[t]
  \centering
 
  \includegraphics[width=1\linewidth]{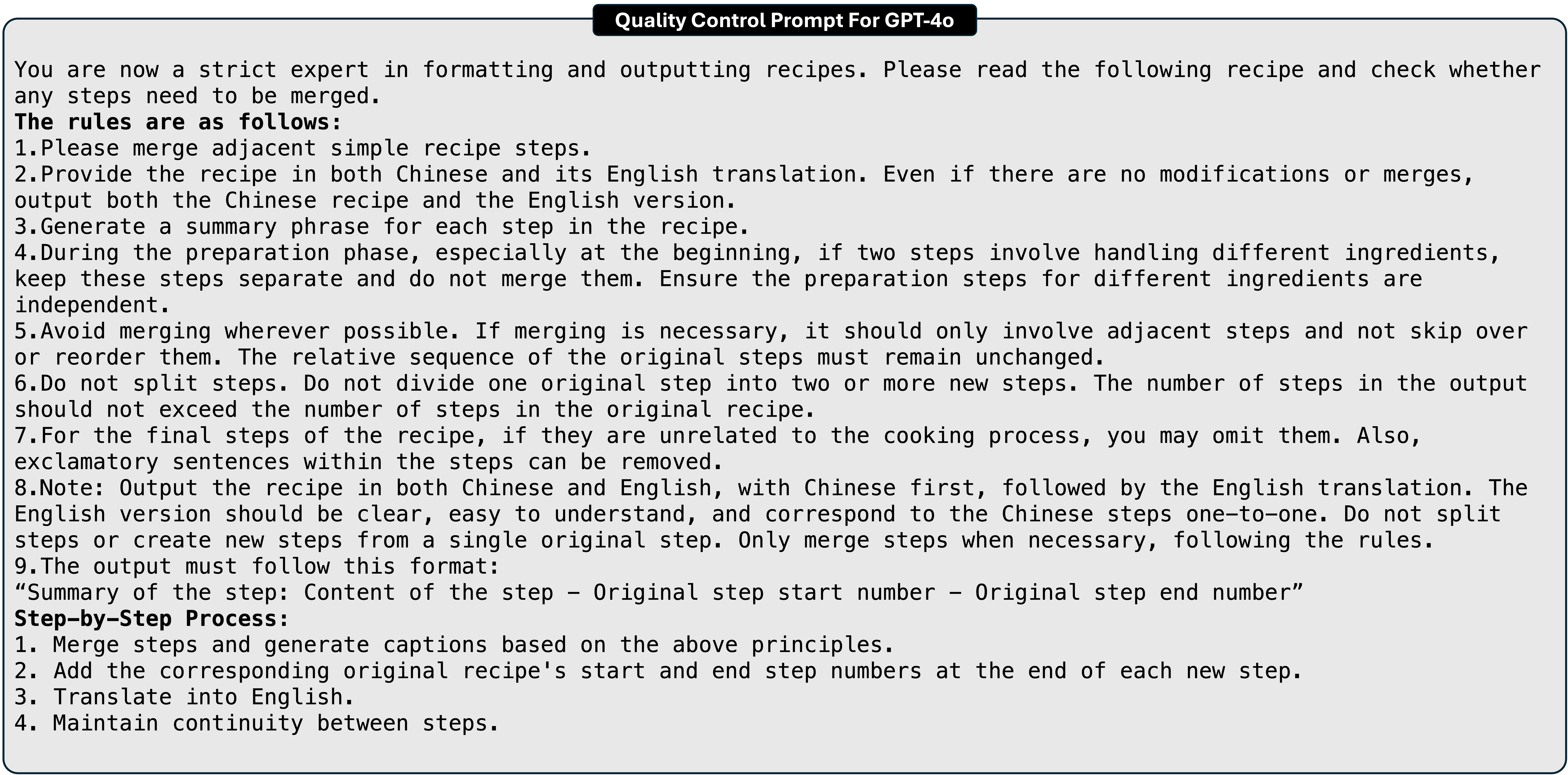} 

  \caption{Quality Control Prompt for GPT-4o. We utilize GPT-4o to merge steps and generate captions.}
  \label{fig:quality_control}
\end{figure*}

\textbf{Prompts for GPT-4o.} To ensure Quality Control, we employ GPT-4o to combine adjacent simple steps, generate captions to prevent over-merging by GPT-4o and translate the output into English. We establish clear principles for GPT-4o and outline a step-by-step process to guide it in producing accurate and appropriate results. The prompt is in Fig. \ref{fig:quality_control}.

\textbf{Human Check.} In addition to using the step faithfulness and goal faithfulness metrics to evaluate the quality of the dataset, we also conduct a human review. Specifically, we ask annotators to verify whether the recipe matched its dish name, whether the steps corresponded to each step image, and whether the images within the same recipe showed continuity. We randomly select 50 recipes for this review. Among these samples, we found one instance where the dish name was mislabeled as ``dumplings" instead of ``steamed buns," and another instance where all the steps lacked specific ingredients. Instead, the textual steps were generic, such as ``as shown in the image" or ``ready to serve".


 


\begin{table*}
  \centering
  \begin{tabular}{@{}p{2cm}p{14cm}@{}}
   \hline
    Type& Keywords \\ 
   \hline
    Regions& Mexican cuisine, Japanese cuisine, Korean cuisine, Indian cuisine, Swiss cuisine, Yunnan cuisine, Su cuisine, Anhui cuisine, Sichuan cuisine, Northeast Chinese cuisine, Zhejiang cuisine, Hubei cuisine, Shandong cuisine, Fujian cuisine, Northwest Chinese cuisine, Shanxi cuisine, Hunan cuisine, Cantonese cuisine, Huaiyang cuisine, Chaozhou cuisine, Xinjiang cuisine, French cuisine, Italian cuisine, German cuisine, Southeast Asian cuisine, Spanish cuisine, Thai cuisine, Chinese cuisine.\\
    \hline
    Dishes &Kung Pao Chicken, Di San Xian, Mapo Tofu, Claypot Rice, Omurice, Boiled Pork Slices, Mashed Potatoes, Braised Pork Belly, Sweet and Sour Pork Tenderloin, Braised Eggplant, Twice-Cooked Pork, Cold Cucumber Salad, Sweet and Sour Pork Ribs, Shredded Pork in Garlic Sauce, Braised Pork Ribs, Spicy Hot Pot, Yellow Braised Chicken, Spicy and Sour Shredded Potatoes, Shredded Pork with Peking Sauce, Steak, Cola Chicken Wings.\\
    \hline
    Staples &Noodles, Cold Noodles, Mixed Noodles, Braised Noodles, Steamed Bun, Dumplings, Steamed Bread, Wonton, Flower Rolls, Pan-Fried Dumplings, Scallion Pancake Roll, Pancake, Scallion Pancake, Stuffed Pancake, Corn Pancake, Stir-Fried Pancakes, Century Egg and Pork Congee, Millet Congee, Seafood Congee, Pumpkin Congee, Mung Bean Congee, Eight Treasure Congee, Rice with Toppings, Mixed Rice, Baked Rice, Braised Rice, Pilaf, Hamburger, Spaghetti, Pizza, Sandwich, Fried Rice, Braised Rice.\\
    \hline
    Ingredients&Cauliflower, Water Spinach, Asparagus, Lettuce Stem, Potato, Celery, Baby Bok Choy, Fennel, Chinese Toon, Rapeseed Greens, Celtuce, Napa Cabbage, Chives, Broccoli, Mustard Greens, Chinese Cabbage, Lettuce, Eggplant, Seaweed, Yellow Chives, Chinese Green Cabbage, Ice Plant, Garlic Sprouts, Vegetable Shoots, Spring Bamboo Shoots, Broccoli, Crown Daisy, Tomato, Spinach, Bitter Chrysanthemum, Cabbage, Enoki Mushroom, King Oyster Mushroom, Shiitake Mushroom, Wood Ear Mushroom, White Shimeji Mushroom, Tremella, Chinese Yam, Carrot, Water Bamboo, White Radish, Sweet Potato, Taro, Lotus Root, Winter Melon, Cucumber, Pumpkin, Loofah, Squash.\\
    \hline
 Dessert   &Toast, Egg Tart, Pudding, Puff, Waffle, Cheesecake, Cream Cake, Layer Cake, Cupcake, Chiffon Cake, Cake Roll, Soufflé, Cream Cheese Cheesecake, Naked Cake, Matcha Cake, Red Velvet Cake, Pound Cake, Strawberry Cake.\\
      \hline
 Others&Breakfast, Cold Dishes, Lunch, Hot Dishes, Midnight Snack, Home-cooked Dishes, Dinner, Side Dishes, Jingxuan.\\
\hline
  \end{tabular}
  \caption{Keywords used in Data Collection. We categorize these 158 keywords into six main groups, including regions and countries, various ingredients, classic dishes,  staple foods, and desserts.}
  \label{tab:keywords}
\end{table*}

\subsection{Keywords used in RecipeGen Benchmark Collection Process}
\label{keywords}
Before gathering recipe data, we first analyze the key characteristics of dishes, based on the idea that a dish is defined by its primary ingredients, cooking techniques, and preparation methods. To ensure a comprehensive collection of recipes, we select a set of 158 keywords encompassing cuisines from various regions, a wide range of cooking methods (e.g., mixing, frying, stir-frying), and categories such as staples, desserts, and ingredient-specific recipes. Additionally, we scrape 996 unique new dishes created by cooking enthusiasts from ``jingxuan" section of the website to test the robustness of the model. Fig. \ref{fig:data_region} shows some distribution of keywords and the full list is provided in Tab. \ref{tab:keywords}. Fig. \ref{fig:example} shows some recipe examples from RecipeGen.

\subsection{Dataset Statistics}

Tab. \ref{tab:RecipeDataset} presents statistical details of our RGB in comparison to existing benchmarks. As shown in Tab. \ref{tab:RecipeDataset}, existing datasets for recipe image generation are limited in the amount of data available, rendering them insufficient for training T2I models. Moreover, the\textbf{ VGSI-Recipe} \cite{yang2021visual}  \footnote{VGSI does not categorize the recipe data, so we search the recipe data in its JSON structure under the ``method" key using the keyword ``cook".} dataset is predominantly Western-focused, offering limited diversity in terms of cooking regions and step counts. Additionally, the dataset includes comic-style images, which diminishes its effectiveness for training models aimed at generating realistic images. On the other hand, while the  \textbf{YouCook2} \cite{zhou2018towards} dataset features a diverse range of cooking styles, it is composed of video data. The data format often obscures essential visual cues due to the position of the chef or suboptimal camera angles, compromising instructional clarity when used for training image-based recipe models. In contrast, RGB offers 21,944 recipes with 139,872 image-step pairs, providing a substantial variety of detailed instructions. Additionally, the images focus solely on the food itself, as they are contributed by amateur food enthusiasts rather than being professionally staged to capture the chef. Tab. \ref{tab:RecipeDataset} further illustrates that most RGB recipes contain over six steps, enhancing the instructional richness of the dataset. Furthermore, to validate quality of the datasets, we assessed goal faithfulness (GF) and step faithfulness (SF) to evaluate the alignment between recipe steps and intended outcomes. GF computes the CLIP score between the caption of the final step and the final image of the recipe. SF computes the CLIP score between each step and its respective image.
As indicated in Tab.  \ref{tab:RecipeDataset}, RGB achieves significantly higher GF and SF scores than VGSI-Recipe, highlighting its superior adherence to recipe objectives and procedural accuracy. 

Overall, our proposed RecipeGen Benchmark (RGB) has four notable features:
\begin{itemize}
    \item \textbf{Broad Step Distribution}: RGB includes recipes with 2 to 25 steps, with an average of 6.4 steps and 55.84\% of recipes containing more than 6 steps, supporting the modeling of long-range semantic relationships.

    \item \textbf{Ingredient Diversity}: RGB provides an average of 9 ingredients per recipe (including seasonings), capturing a rich variety of ingredients and interactions.

    \item \textbf{Variety of Cooking Styles}: The dataset encompasses a wide range of cooking styles enhanced by numerous unique keywords, making it versatile across different culinary processes.

    \item \textbf{Real-World Representativeness}: Collected from various users, RGB consists entirely of real-world recipes and closely resembles the types of instructions that users might upload in practical scenarios. Quality control using GPT-4o ensures that steps are trustworthy and concise, reflecting actual cooking practices.
\end{itemize}

\section{Conclusion}
\label{sec:conclusion}
In this work, we introduced RecipeGen, the first real-world goal-step-image benchmark for recipe image generation, addressing the lack of comprehensive datasets in food computing. We believe that our benchmark can foster further advancements in food computing, particularly in real-world, interactive culinary applications.

\bibliographystyle{IEEEtran}
\bibliography{IEEEabrv,main}

\end{document}